%% file: main.tex
\documentclass[conference]{IEEEtran}
\IEEEoverridecommandlockouts
\usepackage{cite}
\usepackage{amsmath,amssymb,amsfonts}
\usepackage{algorithm}
\usepackage{algpseudocode}
\usepackage{booktabs} % added
\usepackage{multirow}
\usepackage{subcaption}
\usepackage{graphicx}
\usepackage{textcomp}
\usepackage{xcolor}
\usepackage{svg}

\def\BibTeX{{\rm B\kern-.05em{\sc i\kern-.025em b}\kern-.08em
    T\kern-.1667em\lower.7ex\hbox{E}\kern-.125emX}}

\algdef{SE}[DOWHILE]{Do}{doWhile}{\algorithmicdo}[1]{\algorithmicwhile\ #1}
\makeatletter
\newcommand{\linebreakand}{%
  \end{@IEEEauthorhalign}
  \hfill\mbox{}\par
  \mbox{}\hfill\begin{@IEEEauthorhalign}
}
\makeatother
\begin{document}

\title{SafePowerGraph-HIL: Real-Time HIL Validation of Heterogeneous GNNs for Bridging Sim-to-Real Gap in Power Grids \\
\thanks{Corresponsing author: Jun Cao, Email: jun.cao@list.lu. This work was supported partially funded by FNR CORE project LEAP (17042283) and the European Commission under project iSTENTOREHORIZON-CL5-2022-D3-01 with ID-101096787 and We-Forming HORIZON-CL5-2022-D4-02-04 with ID-101123556.}
}
\author{\IEEEauthorblockN{Aoxiang MA}
\IEEEauthorblockA{
\textit{LIST}\\
Esch-Belval, Luxembourg \\
https://orcid.org/0009-0005-9553-5650}
\and
\IEEEauthorblockN{Salah GHAMIZI}
\IEEEauthorblockA{
%\textit{ERIC/ICES} \\
\textit{LIST},~\IEEEmembership{Member,~IEEE}\\
Esch-Belval, Luxembourg \\
https://orcid.org/0000-0002-0738-8250}
\and
\IEEEauthorblockN{Jun CAO}
\IEEEauthorblockA{
\textit{LIST},~\IEEEmembership{Member,~IEEE}\\
Esch-Belval, Luxembourg \\
https://orcid.org/0000-0001-5099-9914}
\linebreakand 
\IEEEauthorblockN{Pedro RODRIGUEZ CORTES}
\IEEEauthorblockA{
\textit{LIST, University of Luxembourg},~\IEEEmembership{Fellow,~IEEE}\\
Esch-Belval, Luxembourg \\
https://orcid.org/0000-0002-1865-0461}
% \author{\IEEEauthorblockN{1st Given Name Surname}
% \IEEEauthorblockA{
% \textit{dept. name of organization (of Aff.)} \\
% https://orcid.org/0000-0000-0000-00000}
% \and
% \IEEEauthorblockN{2nd Given Name Surname}
% \IEEEauthorblockA{
% \textit{dept. name of organization (of Aff.)} \\
% https://orcid.org/0000-0000-0000-00000}
% \and
% \IEEEauthorblockN{3rd Given Name Surname}
% \IEEEauthorblockA{
% \textit{dept. name of organization (of Aff.)} \\
% https://orcid.org/0000-0000-0000-00000}
% \linebreakand 
% \IEEEauthorblockN{4th Given Name Surname}
% \IEEEauthorblockA{
% \textit{dept. name of organization (of Aff.)} \\
% https://orcid.org/0000-0000-0000-00000}

% \and
% \IEEEauthorblockN{Jun CAO}
% \IEEEauthorblockA{
% \textit{LIST},~\IEEEmembership{Member,~IEEE}\\
% Esch-Belval, Luxembourg \\
% https://orcid.org/0000-0001-5099-9914}
% \linebreakand 
% \IEEEauthorblockN{Pedro RODRIGUEZ CORTES}
% \IEEEauthorblockA{
% \textit{LIST, University of Luxembourg},~\IEEEmembership{Fellow,~IEEE}\\
% Esch-Belval, Luxembourg \\
% https://orcid.org/0000-0002-1865-0461}
% \and
% \and
}

\maketitle

\begin{abstract}
\input{pages/0-abstract}
\end{abstract}


\section{Introduction}
\input{pages/1-intro}

\section{Our Approach}
\input{pages/2-method}

\section{Empirical Study}
\input{pages/3-empirical}

\section*{Conclusion and Future Work}
\input{pages/5-conclusion}

\bibliographystyle{unsrt}
\bibliography{bib/main}

\end{document}

%% file: pages/0-abstract.tex
As machine learning (ML) techniques gain prominence in power system research, validating these methods’ effectiveness under real-world conditions requires real-time hardware-in-the-loop (HIL) simulations. HIL simulation platforms enable the integration of computational models with physical devices, allowing rigorous testing across diverse scenarios critical to system resilience and reliability. In this study, we develop a SafePowerGraph-HIL framework that utilizes HIL simulations on the IEEE 9-bus system, modeled in Hypersim, to generate high-fidelity data, which is then transmitted in real-time via SCADA to an AWS cloud database before being input into a Heterogeneous Graph Neural Network (HGNN) model designed for power system state estimation and dynamic analysis. By leveraging Hypersim's capabilities, we simulate complex grid interactions, providing a robust dataset that captures critical parameters for HGNN training. The trained HGNN is subsequently validated using newly generated data under varied system conditions, demonstrating accuracy and robustness in predicting power system states. The results underscore the potential of integrating HIL with advanced neural network architectures to enhance the real-time operational capabilities of power systems. This approach represents a significant advancement toward the development of intelligent, adaptive control strategies that support the robustness and resilience of evolving power grids.

%% file: pages/1-intro.tex
he rapid development of machine learning (ML) has aroused great interest in its potential applications in power system analysis and control. Among ML techniques, Graph Neural Networks (GNNs) have emerged as particularly effective for power systems due to their capacity to model relationships in graph-structured data, such as the topology of electrical grids. Recent studies have demonstrated the potential of GNNs in tasks like fault detection \cite{1}, optimal power flow \cite{2}, and outage prediction \cite{3}, leveraging the inherent structure of power networks. However, while GNNs hold promise for enhancing power system performance, their real-world applicability remains limited by a critical barrier: the gap between simulation-based validation and the unpredictable, dynamic nature of real-world environments, known as the Sim-to-Real Gap.

To bridge this gap, hardware-in-the-loop (HIL) simulation\cite{hil} has emerged as a powerful method for testing and validating data-driven models \cite{4} \cite{8}, like GNNs under realistic conditions. HIL simulation integrates computational models with actual hardware, enabling real-time testing across scenarios that mirror the complex and variable conditions of live power systems. This setup allows for rigorous validation, crucial for verifying the reliability and robustness of ML models in scenarios involving grid disturbances, faults, and other system stresses. 

Several studies have employed HIL simulation to verify machine learning models. For instance, Davis et al.\cite{5} used Bayesian regularized deep neural networks to optimize the charging and discharging control of battery energy storage systems in DC microgrids through HIL simulation, and verified using HIL. In addition, Amir et al. designed a real-time charging controller based on reinforcement learning and verified its effectiveness in bidirectional power flow control of electric vehicles through HIL simulation\cite{6}. The above studies have demonstrated the necessity of hardware verification, but the research on verifying machine learning (ML) models (especially graph neural networks, GNNs) based on HIL platforms is still limited.

In this work, we developed a SafePowerGraph-HIL framework that uses real-time high-fidelity data for fine-tuning and verifying pre-trained Heterogeneous GNNs. The framework uses HIL simulation in Hypersim to verify the accuracy of the GNN model. The data generated by the system in real-time is compared with the output provided by the GNN agent to evaluate its performance in various operational scenarios. Via SCADA, the data is transferred to an AWS cloud database, where it can be integrated and systematically compared with the results of the GNN model. The contributions of our work are threefold, as summarized here.

\begin{itemize}
    \item We propose a SafePowerGraph-HIL framework for real-time hardware-in-the-loop training and validation framework for graph neural networks. We demonstrate that our approach supports state-of-the-art heterogeneous components in power systems.
    \item We build and validate a heterogeneous graph neural network (HGNN) model that achieves high accuracy in state estimation and dynamic analysis under hardware-simulated grid conditions.
    \item We integrate HIL simulation with a cloud database (AWS), providing an innovative solution for real-time monitoring and dynamic response of power system states and enhancing the robustness and adaptability of HGNN in varying system conditions.
\end{itemize}

%% file: pages/2-method.tex
In this section, we describe the SafePowerGraph-HIL verification framework developed to evaluate the dynamics and control accuracy of the power system. The framework is composed of three core components: (1) the SafePowerGraph framework for system modeling and grid simulation.  (2) a real-time hardware-in-the-loop simulation module for generating high-fidelity system data, (3) an AWS cloud database for scalable data storage and processing. Each module is designed to ensure data integrity, efficiency, and precision throughout the framework, facilitating a robust environment to validate AI-driven power system solutions.

We first provide a general overview of our framework, then explain in details each of the three components.
\subsection{SafePowerGraph-HIL Verification Framework Overview}

\begin{figure*}[htbp]
  \centering
  \includegraphics[width=\textwidth]{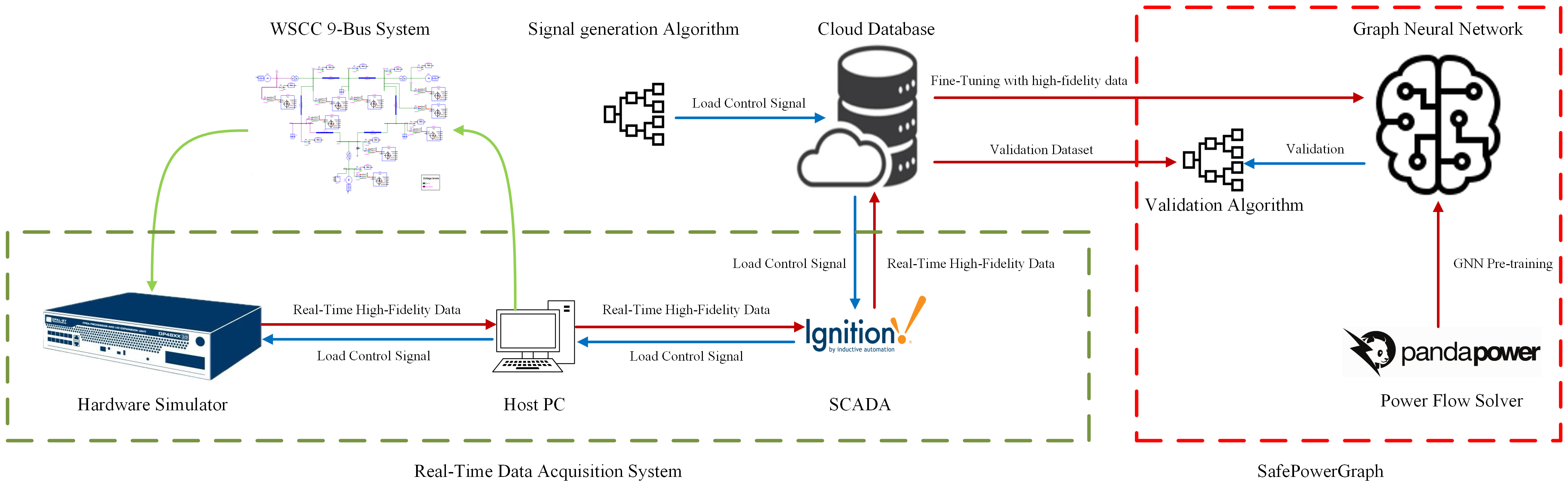} 
  \caption{SafePowerGraph-HIL Verification Framework}
  \label{fig:hil}
  %\vspace{-1em}
\end{figure*}

\subsubsection{Real-Time High-Fidelity Data Generation}
To generate high-fidelity data, a control signal is initiated by SafePowerGraph (using a Python API to the database) to adjust system loads, creating varied operational scenarios (as shown in Fig. \ref{fig:hil}). This signal is temporarily stored in the database, from which the SCADA system retrieves it and applies it to the model to modify load conditions. Concurrently, the power system model within Hypersim executes based on parameters derived from power flow calculations. The real-time system parameters are subsequently transmitted by SCADA to the database for storage, providing a comprehensive data set for GNN training.

\subsubsection{Pre-Trained GNN Validation}
For validating the pre-trained GNN model, a distinct load adjustment control signal (differentiated from training scenarios) is similarly transmitted to the model to capture real-time data. As illustrated in Fig. 1, the newly acquired data are then compared against the predictions of the GNN model, facilitating an evaluation of the model’s performance and accuracy in replicating the dynamic behavior of the system.

\subsection{SafePowerGraph Framework}
\input{pages/2.4-SafePowerGraph}

\subsection{Real-Time Hardware-In-the-Loop Simulation Module}
\subsubsection{Real-time hardware simulator}
\input{pages/2.1-Hardwarehypersim}

\subsubsection{SCADA}
\input{pages/2.2-SCADA}
\subsection{AWS cloud database}
\input{pages/2.3-AWS_database}

%% file: pages/2.4-SafePowerGraph.tex
SafePowerGraph \cite{7} is a framework for building and training graph neural networks to solve PowerFlow and Optimal Powerflow problems using state-of-the-art graph architectures and training paradigms. In SafePowerGraph-HIL, a heterogeneous graph neural network with two GAT graph layers\cite{gnn} is adopted to learn the PF and OPF problem by minimizing three losses:

\textbf{1- A supervised loss} that uses the oracle's solution as a reference and reduces the mean squared error (MSE) between the GNN's predicted output and the oracle's result.

\textbf{2- A constraint-violation loss} that represents component limits as soft constraints, seeking to ensure that outputs adhere to these limits; for example, ensuring the predicted active power of each generator remains within its capacity or maintaining line constraints.

\textbf{3- Complete training loss:} Given the aforementioned, we incorporate the constraints of the problem in the loss of our GNN as regularization terms.
We transform each of the constraints and optimization objectives into a regularization loss.
Given a training graph $\mathcal{G}$, its set of node types $\mathcal{A}$ = \{bus, slack\}, their features $\mathcal{X}$, their ground truth power flow simulation $\mathcal{Y}$ and its predicted output $y$, the training loss function of our GNN (parametrized by $\Theta$) becomes:
\begin{align}
    \mathcal{L}(\Theta) &:= \lambda_{b} \|y_b - \mathcal{Y}_b \|^2_2 + \lambda_{s} \|y_s - \mathcal{Y}_s \|^2_2  \\
    &\quad + \lambda_{b,i} \sum_{i \in \mathcal{A}} \Bigg[\sum_{\omega \in \Omega} \text{ctrloss}(x_i, y_i, \omega) \nonumber \Bigg] 
    \label{eq:losses}
\end{align}

Where the first term is the supervised loss over the outputs of the buses and the slack nodes.
The second term captures the weighted constraint violations of the power grid given the set of constraints $\Omega$ of each type of node.

\paragraph{Dataset generation}
SafePowerGraph mutates the active and reactive powers of individual loads following a uniform distribution centered around the initial configuration, runs the power flow, and generates a synthetic dataset for power flow. The synthetic dataset is a set of samples that each includes the topology of the grid, the parameters of the lines, buses, transformers, generators, and the slack node. The unknown parameters for the power flow problem to be predicted are the active and reactive power of the slack nodes and the voltages and angles of each individual node. 

SafePowerGraph supports communication with SCADA through an SQL database. For each mutated grid, SafePowerGraph sends a new command to the hardware simulated with the new simulation parameters, waits for the simulation to reach a new static state, then fetches the new states of the grid and generates a new training/evaluation sample for the machine learning model. 

\paragraph{Heterogeneous Graph Neural Networks (HGNN) Training Process}
Our approach constructs a Heterogeneous Graph Neural Network (HGNN) based on the framework proposed in \cite{2}, designed to efficiently capture the heterogeneity in power grid data. This architecture has been shown to outperform traditional neural networks and homogeneous GNNs in power flow estimation.

The training process begins by initializing each node $v$ with its feature vector:
\begin{equation}
    h_v^0 = x_{m,v},
\end{equation}
Where $x_{m,v}$ represents the initial attributes specific to each node type (e.g., bus, generator, load). For each layer $k = 1, \dots, K$, the model performs message passing using a Graph Attention Network (GAT) layer, which applies attention weights to the neighbors of each node. The hidden state of node $v$ at layer $k$ is computed as:
\begin{equation}
    h_v^k = \text{ReLU} \left( \sum_{u \in \mathcal{N}(v)} \alpha_{vu}^k W^k h_u^{k-1} \right),
\end{equation}
where $W^k$ denotes the learnable weight matrix for layer $k$, and $\alpha_{vu}^k$ is the attention coefficient that determines the importance of node $u$’s features to node $v$ at layer $k$. These attention coefficients are calculated as:
% \begin{equation}
%     \alpha_{vu}^k = \frac{\exp \left( \text{LeakyReLU} \left( a^k \cdot \left[ W^k h_v^{k-1} \| W^k h_u^{k-1} \right] \right) \right)}{\sum_{j \in \mathcal{N}(v)} \exp \left( \text{LeakyReLU} \left( a^k \cdot \left[ W^k h_v^{k-1} \| W^k h_j^{k-1} \right] \right) \right)},
% \end{equation}
\begin{equation}
    \alpha_{vu}^k = \frac{\exp \left( \sigma \left( a^k \cdot \left[ W^k h_v^{k-1} \| W^k h_u^{k-1} \right] \right) \right)}{\sum_{j \in \mathcal{N}(v)} \exp \left( \sigma \left( a^k \cdot \left[ W^k h_v^{k-1} \| W^k h_j^{k-1} \right] \right) \right)},
\end{equation}

where $\sigma$ represents the LeakyReLU activation function applied to the attention mechanism, $a^k$ is a learnable vector that defines the attention mechanism, and $\|$ denotes concatenation.

To stabilize training, each node’s hidden representation is normalized as follows:
\begin{equation}
    h_v^k \leftarrow \frac{h_v^k}{\| h_v^k \|_2}.
\end{equation}

After processing through $K$ layers, the final hidden states $h_v^K$ are used to produce predictions. For each bus node $b$ and slack node $g$, a linear regression layer outputs the desired quantities:
\begin{equation}
    y_{b} = \text{LINEAR} \left( h_b^K \right), \quad y_{g} = \text{LINEAR} \left( h_g^K \right),
\end{equation}
Where $y_b$ represents the predicted voltage magnitude and angle for bus nodes, and $y_g$ denotes the active and reactive power predictions for slack node.

This layer-wise aggregation and transformation framework, using GAT layers, enables the HGNN to learn complex inter-node relationships with adaptive attention weights, enhancing the model's ability to capture significant patterns within the power grid.

\paragraph{Graph Neural Network Training}

Our evaluation covers two scenarios:

\textbf{(1) GNN performance under distribution shift.} In this first scenario, the GNN are trained only on synthetic datasets simulated with SafePowerGraph, then evaluated using hardware simulator data. In this step, we investigate the generalization capabilities of GNN to unknown noisy hardware sensors.

\textbf{(2) GNN performance with fine-tuning.} In this scenario, we first pre-train the GNN on simulated dataset, then fine-tune with real-time hardware simulation. We evaluate the GNN on the same hardware simulated data as (1). Our study investigates the plasticity of the GNN to hardware environment.

\paragraph{Performance evaluation.}

SafePowerGraph uses hardware simulator data to evaluate the performance of the GNN, in particular, we evaluate the error between the hardware simulated state and the GNN outputs for the bus voltages and angles and the active and reactive powers of the slack node.

%% file: pages/2.1-Hardwarehypersim.tex
This module is responsible for generating high-fidelity data through real-time simulation of power system dynamics. To test the algorithm's functionality, the IEEE 9-bus system model is implemented in the Hypersim simulation environment (see Fig. 1), enabling replication of complex operational scenarios and dynamic responses within the power system. This setup supports real-time adjustments to load sizes, allowing the simulation to mirror various operating conditions.
Within this model, RMS (Root Mean Square) measurement components are employed to capture voltage magnitudes at each bus, while PMU (Phasor Measurement Unit) modules are used to measure phase angles across the buses. Additionally, real-time measurements of active and reactive power are recorded at key nodes, including the slack bus, generators, and loads. These system parameters are subsequently utilized as inputs to the Graph Neural Network (GNN) fine-tuning and validating process, providing a dataset that reflects the real-time state of the system and its dynamic responses under varying conditions.
\begin{figure}[htbp]
  \centering
   \includegraphics[width=\linewidth]{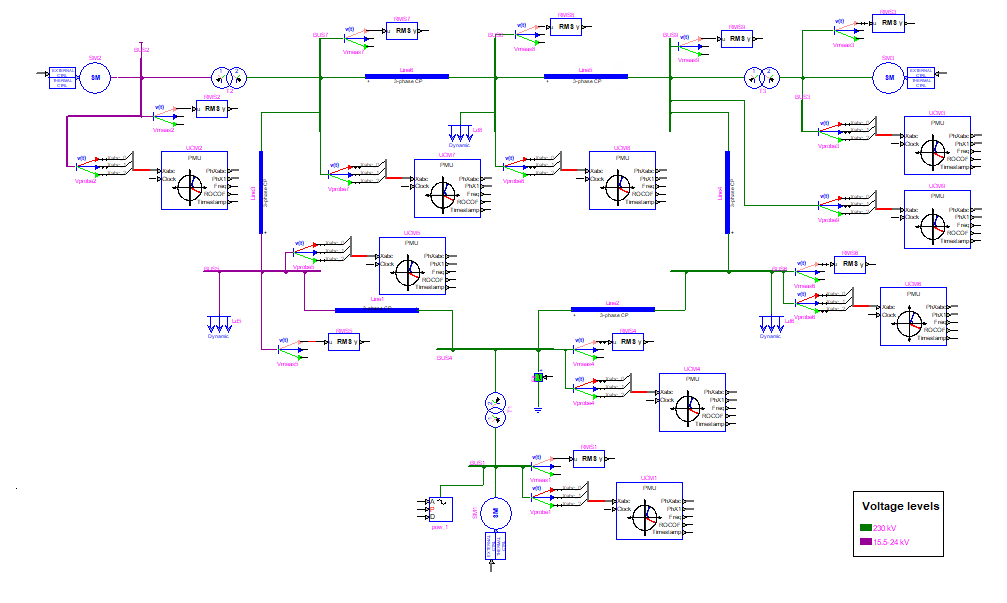} 
   %\includesvg[inkscapelatex=false,width=\linewidth]{figures/to_hetero_bus.svg}
  \caption{WSCC 9-Bus System from HYPERSIM }
  \label{fig:hgnn-graphsage}
  %\vspace{-1em}
\end{figure}

%% file: pages/2.2-SCADA.tex
This module employs the Modbus communication protocol to facilitate real-time data exchange within the simulation model. In this configuration, the hardware simulation model functions as the Modbus slave, while the SCADA system acts as the Modbus master. Designated Modbus tags are configured through an I/O interface, enabling seamless connectivity with the local Ignition SCADA system. Within the Ignition SCADA software, real-time data recording and control signal transmission are achieved through the use of OPC tags and database tags. These tags support the reading and overwriting of values, ensuring continuous synchronization between the hardware model and the database. This setup allows for dynamic updates of load parameters and the real-time tracking of key data points throughout the simulation.

%% file: pages/2.3-AWS_database.tex
We configured a database using AWS RDS infrastructure to allow high scalability and low response time necessary to support the frequency update of the SCADA module. The relational database requires two tables for each case. The first table is used to record the state of the simulated grid at each step; In particular, we record the loads' active and reactive power values, the buses' voltages and angles, the generators' active and reactive powers, and the slack nodes active and reactive powers. Each step is identified by an incremental ID and a unique timestamp. The second table records the orders to the simulation, for instance, the change of load values. This allows a controllable simulation using any external API. The SCADA module monitors any change in this table, and if detected updates the actual HIL simulation parameters accordingly.

%% file: pages/3-empirical.tex
\subsection{Experimental Settings}
\input{pages/3.1-protocol}

\subsection{Results}
\input{pages/4.2-baselines}

%% file: pages/3.1-protocol.tex
\paragraph{Use case}
To validate the effectiveness of our SafePowerGraph-HIL Verification framework, we employed the WSCC 9-Bus system topology. This model serves as a simplified representation of the Western System Coordinating Council (WSCC), comprising 9 buses, 3 generators, and 3 loads. The 9-Bus system provides a foundational platform for simulating essential operational characteristics and dynamic responses of a power system. It offers a controlled environment for assessing the performance of control algorithms and validating model accuracy under varying conditions.

\paragraph{Simulation tool}
\begin{figure}[htbp]
  \centering
   \includegraphics[width=\linewidth]{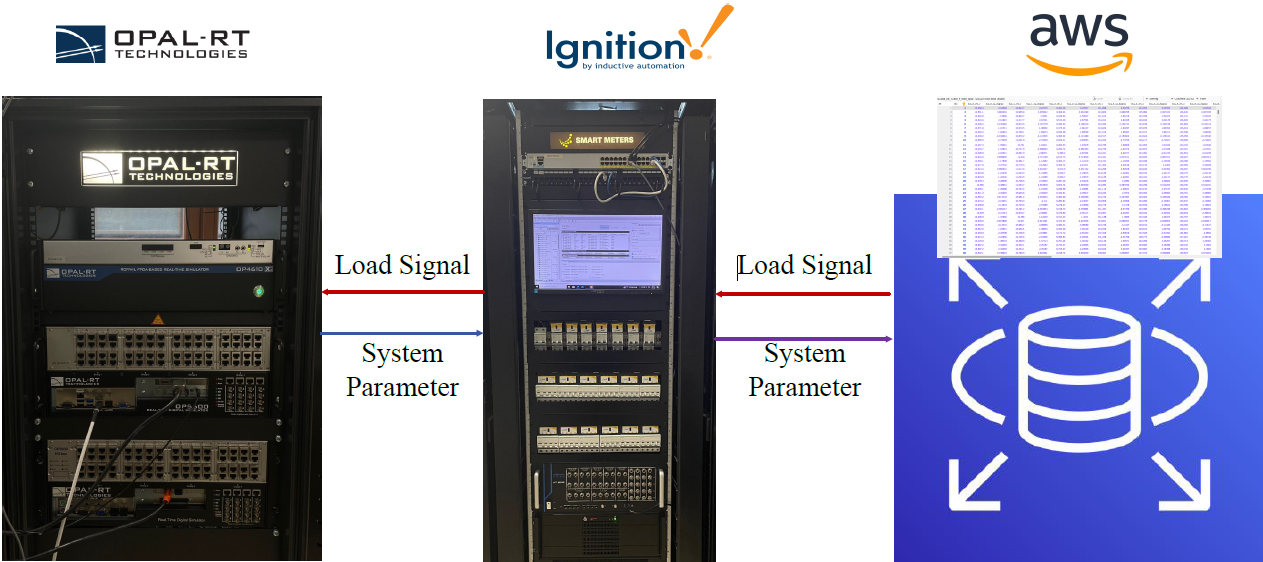} 
  \caption{Our Hardware in the loop infrastructure: Left: Hardware simulator, middle: SCADA interface, right: Real-Time database}
  \label{fig:hgnn-graphsage}
  %\vspace{-1em}
\end{figure}
We use the OPAL4610XG simulator to generate high-fidelity real-time data, connect to the ignition SCADA software through the Modbus communication protocol, obtain new load control signals from the database every 2.5 minutes to input the model, and output model parameters every second to record in the database. For each period of different load levels, we will save a set of data, a total of 500 sets of data for fine-tuning of pre-trained GNN. Figure 3 shows the data generation and storage side hardware experimental component setup.

\paragraph{Realtime database}

We set up a MariaDB database with dynamic scaling. The connection between the database and SCADA is managed by Ignition SQL bridge, and the connection with the python code for training and evaluating the GNNs is handled with the AWS python API.

\paragraph{Neural Network architecture}

Following \cite{2}, we build heterogeneous graph neural networks with two GAT graph layers of 64 neurons each. The outputs of the models are the active and reactive power of the slack nodes and the magnitude and angles of the bus voltages.  

\paragraph{Training and optimization}

We train all the models for 500 epochs and a batch size of 128. We use Adam optimizer and a multistep learning rate, starting at 0.1 and decaying by 0.3 at epochs $\in \{250,375,450\}$. Data augmentation uses a mutation rate of 0.7. The predicted active and reactive powers, the buses' voltages, and angles are trained using an MSE loss.

\paragraph{Metrics}

In our empirical study, we report two metrics for the slack node: normalized squared error for the active and reactive powers obtained through the hardware simulation for the slack node, similarly, we measure the normalized squared error to the bus voltages and angles obtained through the hardware simulation for the buses.

%% file: pages/4.2-baselines.tex
\begin{figure}[htbp]
  \centering
   \includegraphics[width=\linewidth]{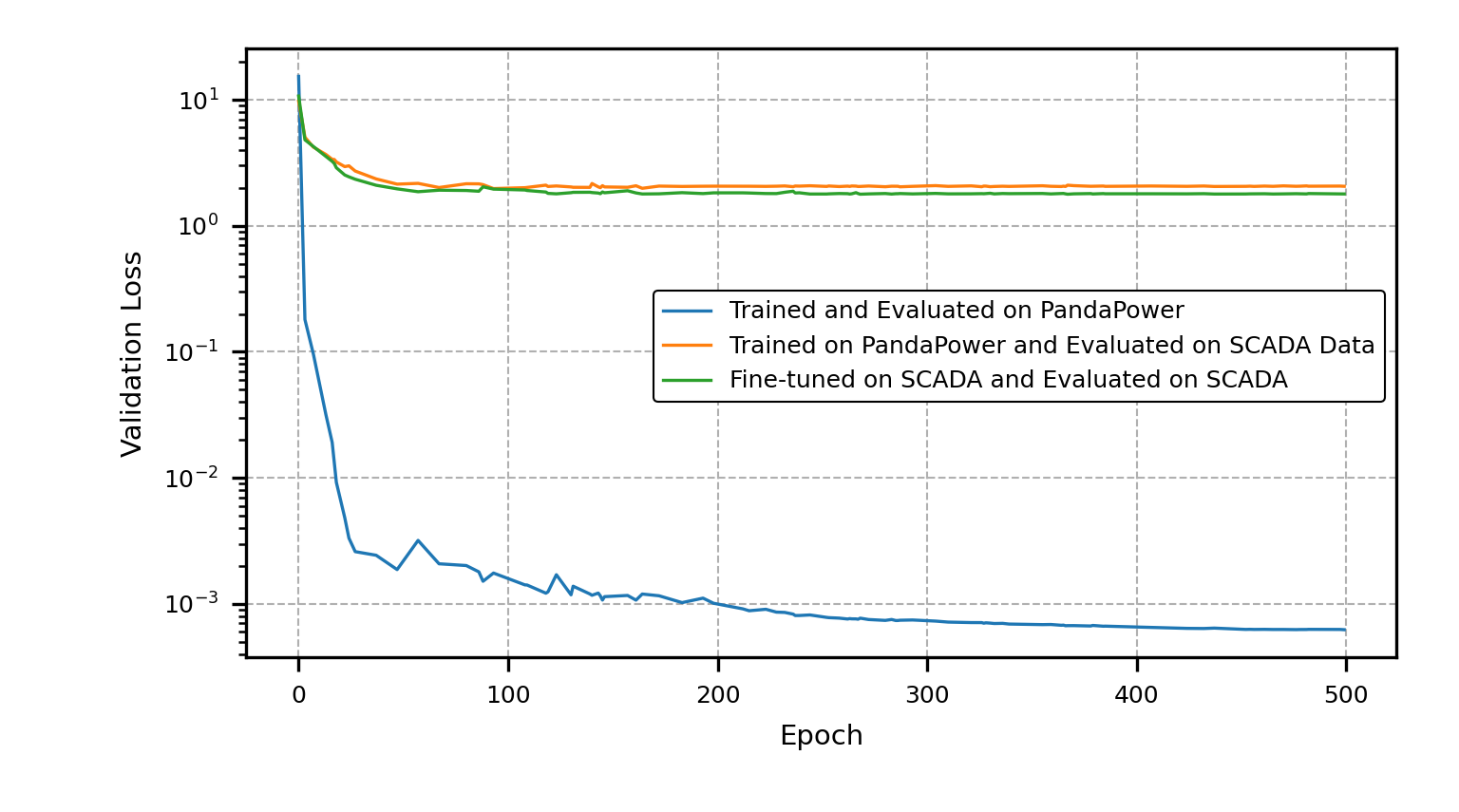} 
  \caption{Validation Loss for Bus Voltage and Angle}
  \label{fig:hgnn-bus}
  \vspace{-1em}
\end{figure}

\begin{figure}[htbp]
  \centering
   \includegraphics[width=\linewidth]{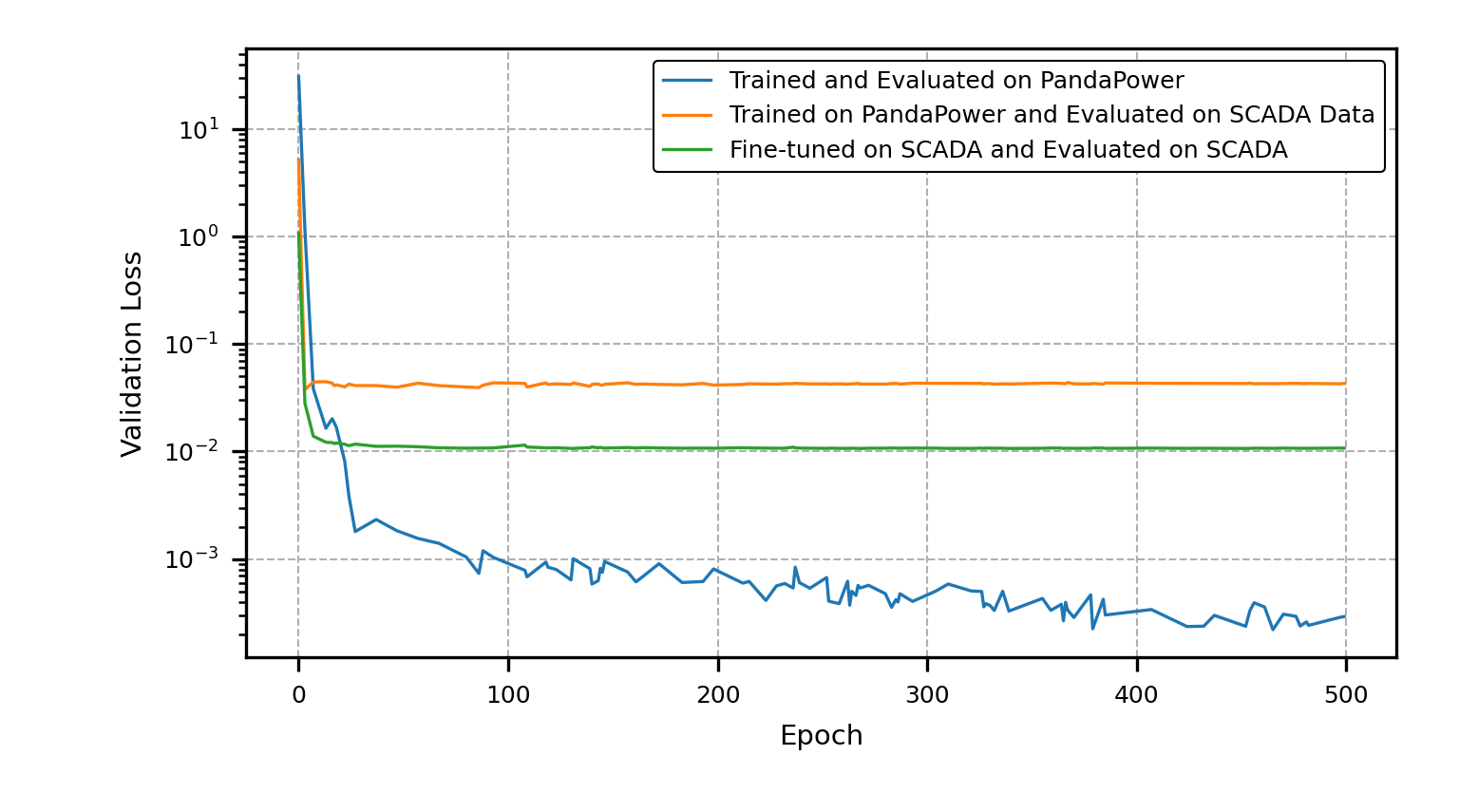} 
  \caption{Validation Loss for Slack Active Power (P) and Reactive Power (Q)}
  \label{fig:hgnn-slack}
  \vspace{-1em}
\end{figure}

Figures \ref{fig:hgnn-bus} and \ref{fig:hgnn-slack} present the validation loss for bus voltage and angle, as well as for slack active power (P) and reactive power (Q), respectively. The model trained and evaluated on PandaPower solver achieves the lowest validation loss, with values of 0.00062 for buses and 0.00029 for the slack node, highlighting the effectiveness of HGNN in the synthetic domain.

However, when the model is evaluated using data generated by the real-time simulator, the validation loss increases significantly to 2.05784 for buses and 0.04302 for the external grid. This substantial degradation underscores the domain shift between synthetic training data and real-world scenarios, thereby limiting the model's applicability under realistic conditions.

To address this issue, the model was fine-tuned using real-time high-fidelity data from SCADA, resulting in notable improvements. The validation loss decreased to 1.78922 for buses, reflecting a 13\% reduction, and to 0.01075 for the external grid, achieving a remarkable 75\% reduction. These results validate the necessity and efficacy of incorporating hardware-in-the-loop (HIL) simulations and real-time data fine-tuning to bridge the gap between simulation and real-world conditions.

%% file: pages/5-conclusion.tex
In this study, we introduced the SafePowerGraph-HIL framework, combining real-time hardware-in-the-loop (HIL) simulations on the IEEE 9-bus system with advanced graph neural networks for power system state estimation and dynamic analysis. Using Hypersim’s high-fidelity data generation and real-time SCADA-to-AWS data transfer, we trained and validated a Heterogeneous Graph Neural Network (HGNN) for accurate state prediction. The framework demonstrated robust performance in capturing complex grid interactions, enhancing PIHGNN’s accuracy and resilience under varied conditions. These findings highlight the potential of integrating HIL with machine learning for real-time power system monitoring and control. Future work will extend the framework to more complex networks with renewable energy sources, exploring HGNN’s adaptability in real-world scenarios and contributing to resilient control strategies for modern power grids.

%% file: main.bbl
\begin{thebibliography}{10}

\bibitem{1}
Jiaxiang Hu, Weihao Hu, Jianjun Chen, Di~Cao, Zhengyuan Zhang, Zhou Liu, Zhe Chen, and Frede Blaabjerg.
\newblock Fault location and classification for distribution systems based on deep graph learning methods.
\newblock {\em Journal of Modern Power Systems and Clean Energy}, 11(1):35--51, 2023.

\bibitem{2}
Salah Ghamizi, Aoxiang Ma, Jun Cao, and Pedro Rodriguez~Cortes.
\newblock Opf-hgnn: Generalizable heterogeneous graph neural networks for ac optimal power flow.
\newblock In {\em 2024 IEEE Power \& Energy Society General Meeting (PESGM)}, pages 1--5, 2024.

\bibitem{3}
Yuzhou Chen, Roshni~Anna Jacob, Yulia~R. Gel, Jie Zhang, and H.~Vincent Poor.
\newblock Learning power grid outages with higher-order topological neural networks.
\newblock {\em IEEE Transactions on Power Systems}, 39(1):720--732, 2024.

\bibitem{hil}
Bethany Sparn, Dheepak Krishnamurthy, Annabelle Pratt, Mark Ruth, and Hongyu Wu.
\newblock Hardware-in-the-loop (hil) simulations for smart grid impact studies.
\newblock In {\em 2018 IEEE Power \& Energy Society General Meeting (PESGM)}, pages 1--5, 2018.

\bibitem{4}
Mohamed Hassouna, Clara Holzhüter, Pawel Lytaev, Josephine Thomas, Bernhard Sick, and Christoph Scholz.
\newblock Graph reinforcement learning for power grids: A comprehensive survey.
\newblock {\em arXiv preprint arXiv: 2407.04522}, 2024.

\bibitem{8}
Songyang Zhang, Tian Liang, Tianshi Cheng, and Venkata Dinavahi.
\newblock Machine learning based modeling for real-time inferencer-in-the-loop hardware emulation of high-speed rail microgrid.
\newblock {\em IEEE Journal of Emerging and Selected Topics in Industrial Electronics}, 3(4):920--932, 2022.

\bibitem{5}
Juan Montoya, Ron Brandl, and et~al. Vishwanath.
\newblock Advanced laboratory testing methods using real-time simulation and hardware-in-the-loop techniques: A survey of smart grid international research facility network activities.
\newblock {\em Energies}, 13(12), 2020.

\bibitem{6}
Pandia~Rajan Jeyaraj, Siva~Prakash Asokan, and Aravind~Chellachi Karthiresan.
\newblock Optimum power flow in dc microgrid employing bayesian regularized deep neural network.
\newblock {\em Electric Power Systems Research}, 205:107730, 2022.

\bibitem{7}
Salah Ghamizi, Aleksandar Bojchevski, Aoxiang Ma, and Jun Cao.
\newblock Safepowergraph: Safety-aware evaluation of graph neural networks for transmission power grids.
\newblock {\em arXiv preprint arXiv:2407.12421}, 2024.

\bibitem{gnn}
William~L. Hamilton, Rex Ying, and Jure Leskovec.
\newblock Inductive representation learning on large graphs.
\newblock In {\em Proceedings of the 31st International Conference on Neural Information Processing Systems}, NIPS'17, page 1025–1035, Red Hook, NY, USA, 2017. Curran Associates Inc.

\end{thebibliography}
